\documentclass{article}

\usepackage{microtype}

\usepackage{hyperref}
\hypersetup{
    colorlinks=true,
    linkcolor=blue,
    filecolor=magenta,
    urlcolor=red,
    citecolor=blue,
}

\usepackage[accepted]{icml2020}
\usepackage{amsmath}
\usepackage{comment}
\usepackage{amsfonts}       
\usepackage{graphicx}
\usepackage{booktabs} 

\usepackage{url}            
\usepackage{subcaption}

\usepackage{nicefrac}       

\icmltitlerunning{Continual Learning from the Perspective of Compression}

\begin{document}

\twocolumn[
\icmltitle{Continual Learning from the Perspective of Compression}



\icmlsetsymbol{equal}{*}

\begin{icmlauthorlist}
\icmlauthor{Xu He}{rug}
\icmlauthor{Min Lin}{mila}
\end{icmlauthorlist}

\icmlaffiliation{rug}{University of Groningen}
\icmlaffiliation{mila}{Mila}

\icmlcorrespondingauthor{Min Lin}{mavenlin@gmail.com}

\icmlkeywords{Continual Learning, MDL, Compression}

\vskip 0.3in
]



\printAffiliationsAndNotice{}  

 \begin{abstract}
 Connectionist models such as neural networks suffer from catastrophic forgetting. In this work, we study this problem from the perspective of information theory and define forgetting as the increase of description lengths of previous data when they are compressed with a sequentially learned model. In addition, we show that continual learning approaches based on variational posterior approximation and generative replay can be considered as approximations to two prequential coding methods in compression, namely, the Bayesian mixture code and maximum likelihood (ML) plug-in code. We compare these approaches in terms of both compression and forgetting and empirically study the reasons that limit the performance of continual learning methods based on variational posterior approximation. To address these limitations, we propose a new continual learning method that combines ML plug-in and Bayesian mixture codes.
 \end{abstract}

\section{Introduction}
\label{sec:intro}
When sequentially learning a non-stationary data stream, neural networks usually suffer from the problem of catastrophic forgetting when the data from the past are not available anymore \citep{mccloskey1989catastrophic}. Recently, this problem has received renewed attention within the field of deep learning, and a flurry of methods (see \citep{parisi2019continual} for a survey) have been proposed to address this issue in order to achieve the goal of continual learning (CL) \citep{ring1998child}.

Intuitively speaking, forgetting can be understood as loss of information about past data. Previous studies on this problem have indirectly measured forgetting by degradation of task-specific performance metrics such as classification accuracy \citep{lopez2017gradient}, regression error \citep{javed2019meta} or the Frechet Inception Distance (FID) \citep{lesort2019generative}. In order to explicitly study the problem of information loss, here we place continual learning within the framework of information theory and Minimum Description Length (MDL) \citep{grunwald2007minimum}. 

According to the MDL principle, any regularity within a dataset can be used for compression, and the best model for a given dataset is the one that results in the minimum \emph{total} description length of the dataset together with the model itself. Therefore, MDL formalizes the Occam's Razor principle for machine learning and provides a criterion for model selection. Other inspirations and foundations for MDL include the ``comprehension is compression" hypothesis by Chaitin \citep{chaitin2002intelligibility} and Solomonoff’s theory of inductive inference \citep{solomonoff1964formal}, which proved that the most likely explanation of some observations of the world is the smallest computer program (in terms of its Kolmogrov Complexity \citep{kolmogorov1968three}) that generates these observations, assuming the world is generated by an unknown computer.

The compression perspective has provided insights in various problems in machine learning, including generalization \citep{arora2018stronger}, unsupervised modeling \citep{ollivier2014auto} and information bottleneck \citep{tishby2015deep}. Recently, \citep{blier2018description} showed that one particular coding method from the MDL toolbox called \emph{prequential coding} outperforms the other coding methods at explaining why deep learning models generalize well despite their large number of parameters, displaying strong correlation between generalization and compression.

In this work, we show that the compression perspective can also shed light on continual learning. Our contributions are:\\
\begin{itemize}
    \item We provide a formal definition of forgetting based on information theory, which allows us to evaluate continual learning methods for generative models in a principled way.
    \item We show that two major paradigms of continual learning (variational continual learning and generative replay) can be interpreted as approximations of two prequential coding methods (Bayesian mixture coding and ML plug-in coding) in the MDL framework, thus establishing a connection between continual learning and compression.
    \item We compare different CL methods for continual generative modeling using the proposed definition of forgetting and empirically study the limitations of the variational continual learning approaches.
    \item We introduce a new CL method that combines the prediction strategies of the two prequential coding methods and show that it improves over the variational continual learning.
    
\end{itemize}

\section{Minimum Description Length}
\label{sec:MDL}
 Formally, given a dataset $x^n:=\{x_1, \dots, x_n\}$ where each $x_i$ is from a space of observation $\mathcal{X}$, the overall description length of $x^n$ under the model with parameters $\theta$ can be separated into two parts:
\begin{equation}
\label{eqn:two_part}
    L(x^n)=L(x^n|\theta)+L(\theta)
\end{equation}
The first part $L(x^n|\theta)$ describes the codelength (number of bits) of $x^n$ when compressed with the help of a model parametrized  by $\theta$, and the second part $L(\theta)$ corresponds to the codelength of the model itself. A model with more capacity can fit the data better, thus resulting in shorter codelength of the first part. However, higher capacity might also lead to longer description length of the model itself. Therefore, the best explanation for $x^n$, according to MDL, is given by the model that minimizes the total length $L(x^n)$. For probabilistic models, the code length $L(x^n|\theta)$ can be computed using the Shannon-Fano code.

\paragraph{Shannon-Fano Code}
Let $p(x|\theta)$ be a probability density (or mass) function over $\mathcal{X}$ determined by parameters $\theta$, then for any $x\in \mathcal{X}$, there exists a prefix code that losslessly compresses $x$ with code length
\begin{equation}
\label{eqn:shannon}
    L(x|\theta) =  -\log_2 p(x|\theta)
\end{equation}
In this work, we use the codelength function $L$ only as a metric to measure the amount of information and do not care about the actual encoding of data, hence we adopt the idealized view of code lengths which does not require them to be integers. 
\paragraph{Bayesian Code}
One way to encode the parameters $\theta$ is to consider these parameters also as random variables and use the following Bayesian coding scheme:
\begin{align}
    \log p(x^n) &= \log \int_\theta p(x^n|\theta)p(\theta) \\
    &\geq \int_\theta q(\theta|x^n)\log p(x^n|\theta)d\theta - D_{\mathrm{KL}}\left[q(\theta|x^n)\|p(\theta)\right]
    \label{eqn:elbo}
\end{align}
where (\ref{eqn:elbo}) is the evidence lower bound (ELBO) used in variational Bayesian methods. The equality holds when  $q(\theta|x^n)=p(\theta|x^n)$.
The first term corresponds to $L(x^n|\theta)$, describing the expected code length of $x^n$ if encoded with $\theta$ sampled from the approximate posterior $q(\theta|x^n)$. The second term corresponds to $L(\theta)$, describing the code length of $\theta$ using the bits-back coding \citep{hinton1993keeping} with a fixed prior distribution $p(\theta)$.

\paragraph{Prequential (Online) Code}
Another way to compute the total description length $L(x^n)$ without explicitly separating $L(x^n)$ into two parts is to use the prequential or online code, which considers a dataset $x^n$ as a sequence of observations and encode each observation $x_t$ at time $t$ by predicting its value based on past observations $x^{t-1}$. Note that for any $x^t$, we can write its Shannon-Fano code as a sum of negative log conditional probabilities:
\begin{align}
\label{eqn:prequential}
    -\log p(x^n) &= \sum_{t=1}^n-\log p(x_t|x^{t-1})
\end{align}
Hence the total description length can be computed by accumulating the code length $- \log p(x_t|x^{t-1})$ at every step $t$ using a prediction strategy. Formally, a prediction strategy is a function $S:\bigcup_{0\leq t\leq n} \mathcal{X}^t\to\mathcal{P}_\mathcal{X}$ that maps any initial observations $x^{t-1}$ to a distribution on $\mathcal{X}$ corresponding to the conditional model $p(X_t|x^{t-1})$. For example, \emph{Bayesian mixture code} uses the following prediction strategy:
\begin{align}
\label{eqn:bayes_code}
    p(X_t|x^{t-1})=\int_\theta p(X_t|\theta)p(\theta|x^{t-1})
\end{align}
where $p(\theta|x^{t-1})$ is the posterior of model parameters given all previous data $x^{t-1}$.\\
Another example of prequential coding is the \emph{maximum likelihood (ML) plug-in code}, which encodes each observation by
\begin{align}
\label{eqn:ml_plugin}
    p(X_t|x^{t-1})&=p(X_t|\theta(x^{t-1}))
\end{align}
where $\theta(x^{t-1})$ is given by a maximum likelihood estimator:
\begin{align}
\label{eqn:ml}
    \theta(x^{t-1})=\arg\max_\theta \log p(x^{t-1}|\theta)
\end{align}

As we will show in the next section, this online incremental view of prequential coding is well-aligned with the setting of continual learning.

\section{Continual Learning}
Continual learning studies the scenario where data are provided at different time in a sequential manner, and the learning algorithm incrementally updates the model parameters as more data are presented. However, due to reasons such as privacy or limited memory budget, the learning algorithm may not always store all the data $x^{t-1}$ it has seen so far. Alternatively, even if all data are stored, naively retraining on all previous data will increase the run time per step over time, which may not be affordable for long data streams. On the other hand, if the model is only trained on the data presented recently, it usually suffers from ``catastrophic forgetting''. This problem is especially severe when the data stream is non-stationary. Here we provide a definition of forgetting that allows us to precisely measure the amount of information forgotten.
\subsection{An Information-Theoretic Definition of Forgetting}
We can see from (\ref{eqn:two_part}) that when the data $x^n$ is compressed with the help of a model, the information of $x^n$ is divided into two parts, with one part contained in the code words of $x^n$ and the other part contained in the model. Hence, the model alone is not enough to recover the data losslessly, it requires also the code words of the compressed data, which can be understood as the information of $x^n$ not captured by the model. This intuition leads to our definition of forgetting: for any data $x$, the amount of forgetting about $x$ caused by changing the model parameters from $\theta_i$ to $\theta_j$ is given as the increase of the codelength of $x$:
\begin{align}
\label{eqn:forgetting}
    \mathcal{F}[\theta_j/\theta_i](x):=\max (L(x|\theta_j)-L(x|\theta_i), 0)
\end{align}
In other words, the number of extra bits required for the model to losslessly recover $x$ is the amount of information the model has forgotten about $x$. The $\max(\cdot, 0)$ function in (\ref{eqn:forgetting}) is used to prevent ``negative forgetting", if the codelength does not increase, there is no forgetting.

In the prequential setting, at any time $t$, we can measure the \emph{cumulative average forgetting} of a learning algorithm over the entire sequence $x^t$ seen so far by the following metric:
\begin{align}
\label{eqn:accu_forget}
    \mathcal{C}(x^t):=&\frac{1}{t}\sum_{i=1}^t\mathcal{F}[\theta_t/\theta_i](x_i)\\
    =&\frac{1}{t}\sum_{i=1}^{t} \max(L(x_i|\theta_t)-L(x_i|\theta_i), 0)
\end{align}
where $\theta_i$ are the parameters given by the learning algorithm at time $i$, after seeing $x^i$.
\subsection{Prequential Interpretation of CL Approaches}
A primary goal of continual learning is to overcome or to alleviate catastrophic forgetting. Many CL methods have been proposed for this purpose \citep{parisi2019continual}. Some common paradigms emerged from these methods. One paradigm is based on the sequential nature of Bayesian inference and it applies variational methods to approximating the parameter posterior. The approximated posterior is then used as the prior for future learning. Examples of this paradigm include Bayesian Online Learning \citep{opper1998bayesian}, Variational Continual Learning (VCL) \citep{2018variational} and Online Structured Laplace Approximations \citep{ritter2018online}. Some regularization-based CL methods such as Elastic Weight Consolidation (EWC) \cite{kirkpatrick2017overcoming} and Uncertainty-guided Continual Bayesian Neural Networks (UCB) \citep{Ebrahimi2020Uncertainty-guided} are also motivated by this idea. Another paradigm combats forgetting by training a generative model of past data. When new data are provided later, the samples from the generative model are replayed instead of the original data. Representative works of this class are Deep Generative Replay (DGR) \citep{shin2017continual}, Memory Replay GAN \citep{NIPS2018_7836} and Continual Unsupervised Representation Learning (CURL) \citep{rao2019continual}.  \citet{farquhar2019unifying} proposed a unifying Bayesian view that encompasses both paradigms. From the perspective of MDL, these two types of approaches can be seen as approximations of the two prequential coding methods introduced before. We discuss their relationships below using VCL and Generative Replay as examples.

\paragraph{VCL} The first paradigm corresponds to the Bayesian mixture code, which uses (\ref{eqn:bayes_code}) as its prediction strategy. 
Since the exact posterior $p(\theta|x^{t-1})$ in (\ref{eqn:bayes_code}) depends on all previous data $x^{t-1}$, which might not always be available in the continual learning scenario, VCL therefore uses a variational model $q(\theta|x^{t-1})$ to approximate the real posterior $p(\theta|x^{t-1})$. By the Bayes' rule, we have 
\begin{align}
\label{eqn:posterior}
   p(\theta|x^{t})=\frac{p(x_t|\theta)p(\theta|x^{t-1})}{Z_{t}}
\end{align}
where $Z_t:=\int_\theta p(x_t|\theta)p(\theta|x^{t-1})$ is a normalizing factor that does not depend on $\theta$. Therefore, $q(\theta|x^{t})$ can be updated recursively by minimizing the following KL divergence at every step $t$:  
\begin{align}
\label{eqn:vcl}
   q(\theta|x^t)=\arg\min_{q(\theta)} D_{\text{KL}}(q(\theta)||\frac{1}{Z_t}p(x_t|\theta)q(\theta|x^{t-1}))
\end{align}
The first approximate posterior $q(\theta|x^0)$ is usually a fixed prior chosen before learning starts: $q(\theta|x^0)=p(\theta)$. Since $Z_t$ does not depend on $\theta$, it is not required for the optimization above.

\paragraph{Replay} The second paradigm, on the other hand, corresponds to the ML plug-in code, which uses the prediction strategy defined in (\ref{eqn:ml_plugin}) and (\ref{eqn:ml}). The objective function in (\ref{eqn:ml}) requires all previous data $x^{t-1}$. In order to relax this requirement for continual learning, replay-based methods approximate the real objective function using a generative model $p(x|\theta)$. Assuming $x_i\in x^{t}$ are independent given $\theta$, we have:
\begin{align}
\label{eqn:DGR_loss}
    \log p(x^{t}|\theta)=&\sum_{i=1}^{t}\log p(x_i|\theta)=\log p(x_{t}|\theta)+\sum_{i=1}^{t-1}\log p(x_i|\theta) \nonumber\\
    \approx& \log p(x_{t}|\theta) + (t-1)\mathbb{E}_{p(x|\theta_{t-1}))}[\log p(x|\theta)] \nonumber\\
    =& \log p(x_{t-1}|\theta) + (t-1)\int_x p(x|\theta_{t-1})\log p(x|\theta)
\end{align}
where $\theta_{t-1}$ are the parameters of the generative model at time $t-1$ and the expected value can be computed by Monte Carlo sampling. In this way the generative model can be recursively updated by the following rule:
\begin{align}
\label{eqn:DGR_update}
    \theta_{t} = \arg\max_\theta \log p(x_{t}|\theta)+(t-1)\int_x p(x|\theta_{t-1})\log p(x|\theta)
\end{align}

\paragraph{Measuring Forgetting and Compression}
We empirically compare the two prequential coding methods and their continual learning approximations in terms of both forgetting and compression on the MNIST dataset. Data are presented in a class-incremental fashion: first only images of digit zero, then only images of digit one, and so on. Throughout the paper, we use a variational autoencoder (VAE)\citep{kingma2013auto} for encoding and generative modelling. In this experiment, one observation corresponds to an entire class of images. 

\begin{figure}[t!]
  \centering
  \includegraphics[width=\columnwidth]{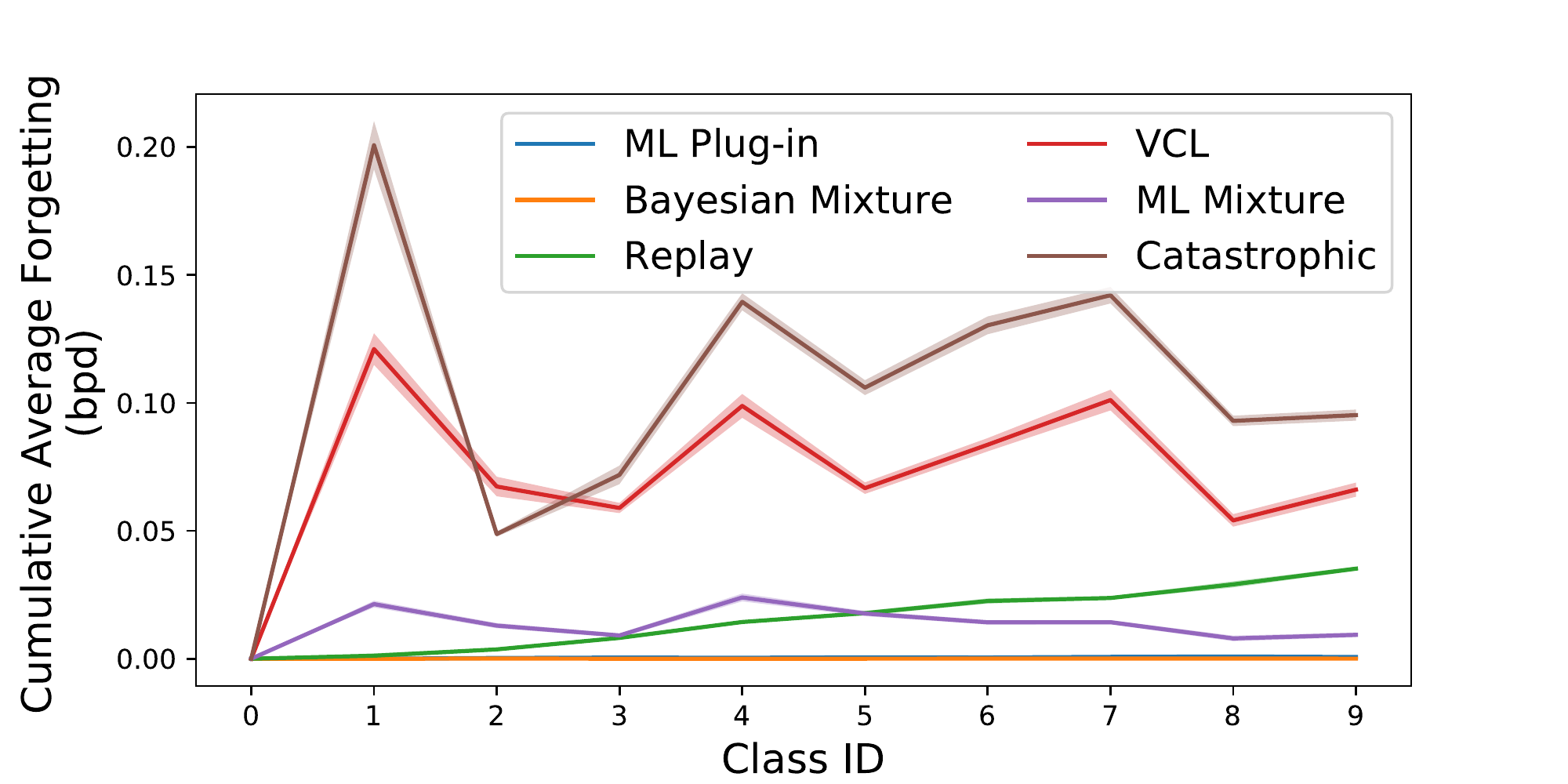}
  \caption{Cumulative average forgetting (defined in (\ref{eqn:accu_forget})) measured after learning each digit in MNIST. The resulting codelength was divided by the number of pixels in an image to get bits per dimension (bpd). The means and standard errors of the mean (SEM) were computed over ten random mini-batches of 32 images from each class. The shaded region correspond to 3 times of SEM. Here ``Catastrophic'' corresponds to the method that trains the model only on images from the current class. The ``ML Mixture'' method is introduced in Sec. \ref{sec:mlmixture}. Both ML plug-in and Bayesian Mixture have near zero forgetting, so their curves are overlapping with each other.}
  \label{forgetting}
\end{figure}

Fig.\ref{forgetting} shows the cumulative average forgetting  after learning each digit. Since ML plug-in and Bayesian Mixture methods are trained on all data seen so far, they have almost zero forgetting. VCL and Replay, on the other hand, do not have access to previous data, so information are lost over time. However, thanks to the variational posterior and generated samples, they achieve less forgetting than the Catastrophic baseline, which only trains the model on images of the current digit.

The results on prequential coding as defined in (\ref{eqn:prequential}) are shown in Fig. \ref{compression}. After all the images of one class are presented, the model is updated and used to encode all the images of the next class. The left plot shows the codelength of each class, since models are only trained on the images from previous classes, the results shown here correspond to the concept of forward transfer in continual learning. We can see that Bayesian Mixture and VCL have high codelength at the beginning, when there is no forgetting yet. The reason for this phenomenon is discussed in Sec. \ref{sec:overhead}. The plot on the right shows the prequential codelength of the entire MNIST dataset measured in bits per dimension.

\section{Limitations of VCL}
 \begin{figure}[h!]
	\begin{subfigure}[b]{\columnwidth}
	 \centering
		\includegraphics[width=0.9\textwidth]{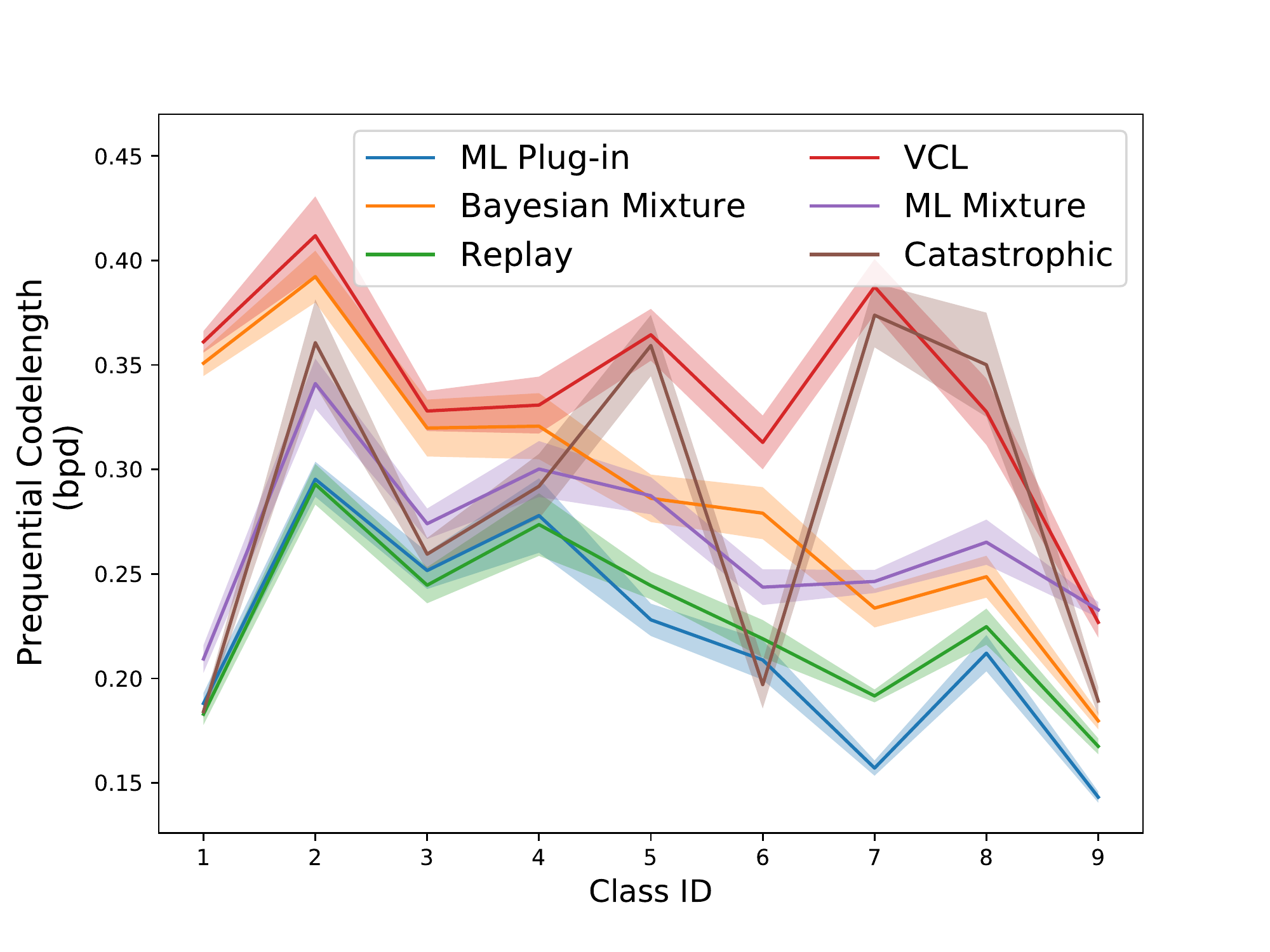}
		\caption{Prequential codelength of images in class $i$, measured after the model has seen all images in class $i-1$. Since the images of the first class are encoded with a predefined model for all methods, their codelength does not affect the comparison results.}
    \end{subfigure}
    \begin{subfigure}[b]{\columnwidth}
        \centering
		\includegraphics[width=0.9\textwidth]{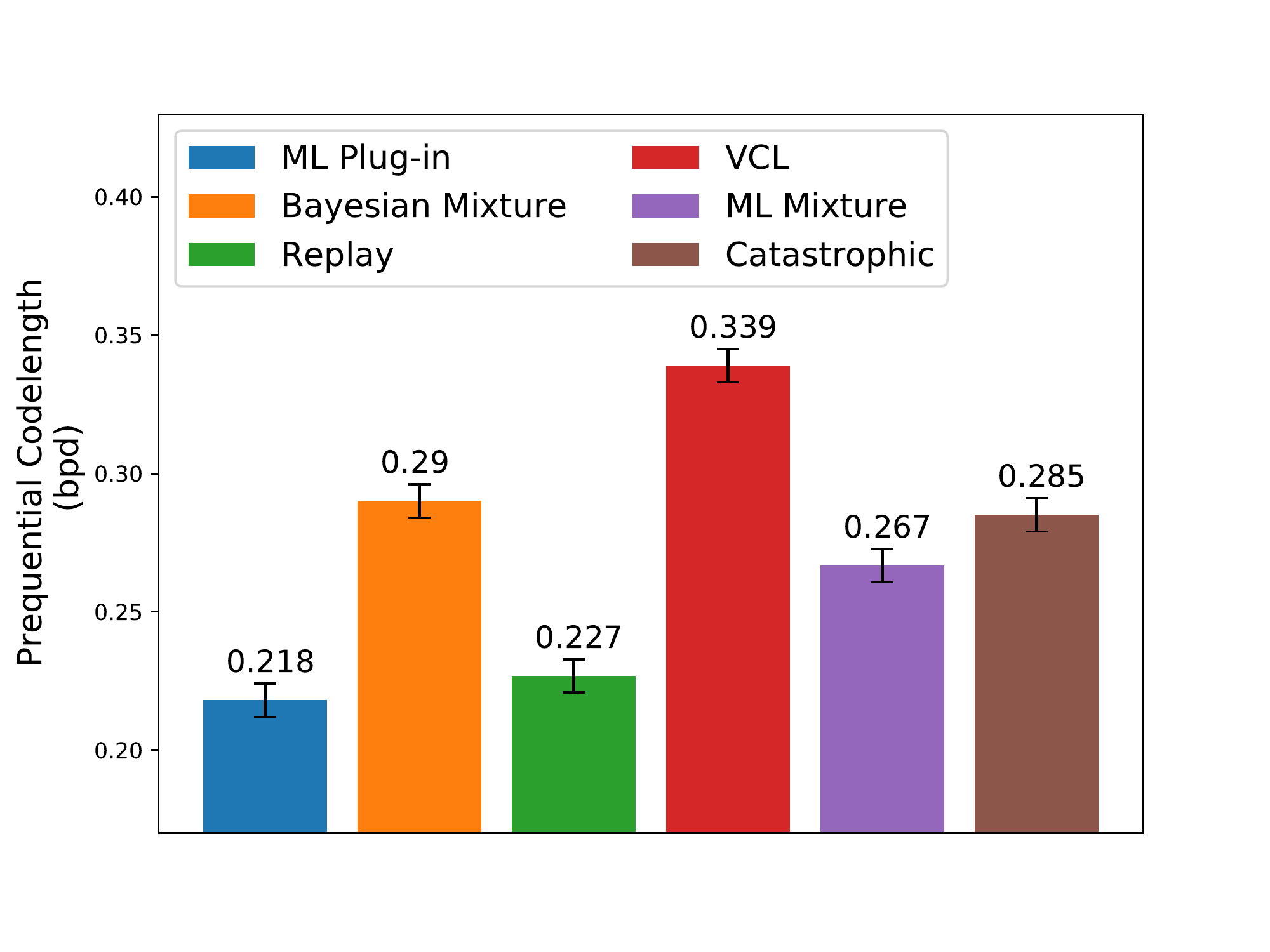}
		\caption{The total prequential codelength of the MNIST dataset, averaged over number of samples.}
    \end{subfigure}
	\caption{Compression performance of different coding methods on the MNIST dataset. The shaded region correspond to 3 times of SEM.}		\label{compression}
\end{figure}

It can be seen from both Fig.\ref{forgetting} and \ref{compression}, VCL performs poorly compared to generative replay. A systematic comparison between VCL  and replay on discriminative tasks was carried out by \citet{farquhar2018towards}. VCL was reported to work similarly to generative replay on simple tasks like Permuted MNIST. On the more challenging Split MNIST task, VCL requires both a multi-head output layer and a coreset of past examples in order to be comparable with generative replay. For generative modeling, the results in Fig. \ref{forgetting} confirm the limitation of VCL at retaining information of past data. In this section, we analyze the reasons that limit the performance of VCL.

\subsection{Overhead of Bayesian Mixture due to Prior}
\label{sec:overhead}
\begin{figure}[b!]
  \centering
  \includegraphics[width=\columnwidth]{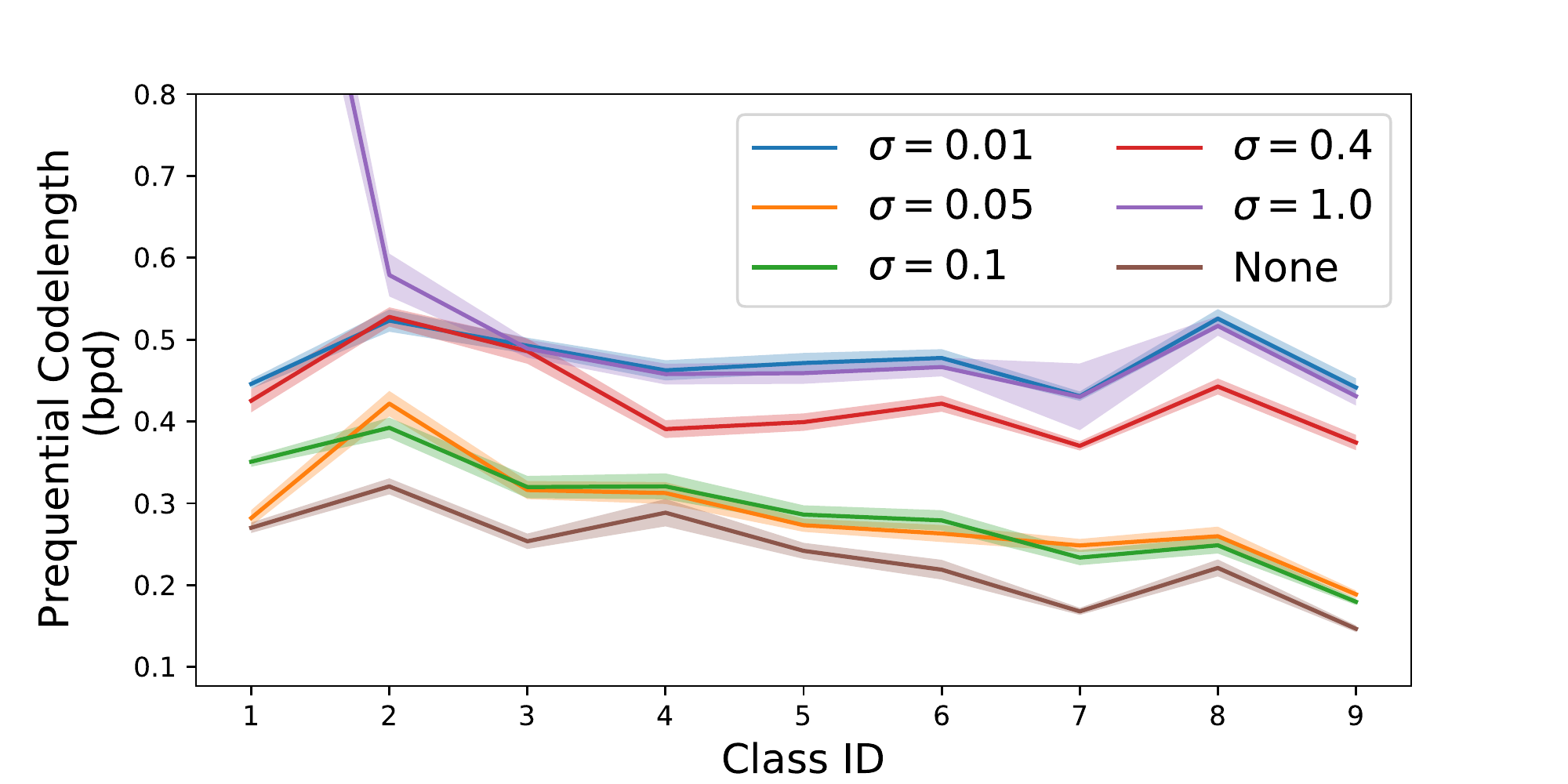}
  \caption{Prequential description length of Bayesian Mixture code using Gaussian priors with different standard deviations $\sigma$. Here, ``None'' means no prior is used, and the KL term in (\ref{eqn:elbo}) is not optimized.}
  \label{prior}
\end{figure}
With deep generative models, the integration in Bayesian mixture code defined in (\ref{eqn:bayes_code}) is intractable because $p(x|\theta)$ is a neural network generator. Therefore, the ELBO is used in practice as an approximation, which adapts $q(\theta)$ in (\ref{eqn:elbo}) to minimize the total description length. We make two observations here. First, in VCL, the best achievable description length is predetermined by the selected prior $p(\theta)$. Second, compared to ML plug-in code, which directly minimizes the encoding length $L(x|\theta)$, the approximate Bayesian code has an extra KL regularization, indicating a longer description length $L(x|\theta)$ for the code words.   Furthermore, since we do not explicitly encode data and keep their code words in continual learning, a larger encoding length means more information is lost about the data.

This overhead is negligible asymptotically if the data sequence is very long. This can be seen from the relative weights of the two terms in (\ref{eqn:elbo}). The relative weight of the KL regularization diminishes when the data sequence grows longer. This is why we see a high initial codelength for Bayesian Mixture in Fig. \ref{compression} (a), but gradually it closes the gap to ML plug-in.

However, for a sequence with fixed length, the overhead due to prior can be significant and is very sensitive to the choice of the prior. Figure \ref{prior} compares the prequential codelength of Bayesian Mixture code using Gaussion priors with different standard deviation $\sigma$. One can see that good performance can be achieved only when $\sigma$ is within a small region around $0.1$. And the best description length is obtained when the prior is not used. Similar effects were observed for VCL in \citep{swaroop2019improving}, where the authors showed the choice of prior variance has great impact on the performance of VCL.

\subsection{The Variational Gap}
\label{sec:var_gap}
As we describe before, in continual learning, the true posterior $p(\theta|x^{t-1})$ is not available, thus VCL approximates it with a variational posterior $q(\theta|x^{t-1})$. The underlying assumption is that when their KL divergence is small enough: $D_{\mathrm{KL}}[q(\theta|x^{t-1})\|p(\theta|x^{t-1})]<\epsilon$, the following approximations are good enough for continual learning:
\begin{align}
    \label{eqn:vcl_approximation}
    &q(\theta|x^{t-1})\approx p(\theta|x^{t-1})\\
    \label{eqn:real_posterior}
    \mathcal{L}^{\mathrm{VCL}}_t(q)=&\int_\theta q(\theta|x^t)\log p(X_t|\theta) - D_{\mathrm{KL}}[q(\theta|x^t)\|p(\theta|x^{t-1})]\\
    \label{eqn:approx_posterior}
    \approx&\int_\theta q(\theta|x^t)\log p(X_t|\theta) - D_{\mathrm{KL}}[q(\theta|x^t)\|q(\theta|x^{t-1})]
\end{align}

However, for VCL with a simple distribution family such as diagonal Gaussian, this assumption can hardly hold, as the real posterior is proportional to $p(x^{t-1}|\theta)p(\theta)$, which is usually complex and multimodal. Since the only difference between VCL and Bayesian Mixture code is the substitution of (\ref{eqn:real_posterior}) by (\ref{eqn:approx_posterior}), the error between these two terms is entirely responsible for their large performance gap observed in Fig.\ref{forgetting} and \ref{compression}. One potential remedy to this problem is to use more complex distributions on the parameter space, for example, normalizing flow as a Bayesian hypernetwork \citep{krueger2017bayesian}. However, unlike Gaussian distributions, the KL divergence of complex distributions usually cannot be calculated analytically. It is also unclear whether sampling in the parameter space to minimize $D_{\mathrm{KL}}[q(\theta|x^{t-1})\|p(\theta|x^{t-1})]$ would be any cheaper than sampling in the data space for replay.

Furthermore, the approximation error between (\ref{eqn:real_posterior}) and (\ref{eqn:approx_posterior}) could be arbitrarily large even if $D_{\mathrm{KL}}[q(\theta|x^{t-1})\|p(\theta|x^{t-1})]<\epsilon$, which only states that $\log\frac{q(\theta|x^{t-1})}{p(\theta|x^{t-1})}$ is small in its expectation over the distribution $q(\theta|x^{t-1})$. As shown in (\ref{eqn:approx_err}), the approximation error is the expectation of the same log ratio over $q(\theta|x^{t-1})p(x_t|\theta)/Z_t$. Unfortunately, this distribution could be very different from $q(\theta|x^{t-1})$. It depends on $p(x_t|\theta)$, which is not available at the time $q(\theta|x^{t-1})$ is solved.  One sufficient condition to bound the substitution error is $\max_\theta|\log\frac{q(\theta|x^{t-1})}{p(\theta|x^{t-1})}|<\epsilon$. However, this involves an minimax optimization of the log ratio, which is a hard problem itself.

\begin{align}
\label{eqn:approx_err}
    \Delta(\mathcal{L}^{\mathrm{VCL}}_t)&=-\int_\theta q(\theta|x^{t})\log\frac{q(\theta|x^{t-1})}{p(\theta|x^{t-1})} \nonumber \\
    &=-\int_\theta \frac{1}{Z_t}q(\theta|x^{t-1})p(x_t|\theta) \log\frac{q(\theta|x^{t-1})}{p(\theta|x^{t-1})}
\end{align}

\section{ML Mixture Code}
\label{sec:mlmixture}
In order to overcome the two limitations discussed above, we can combine the prediction strategies of Bayesian mixture and ML plug-in using a hierarchical model with hyperparameters $\phi$. For diagonal Gaussian distribution, $\phi$ stands for the mean and variance of $\theta$. The resulting prediction strategy, which we call \emph{maximum likelihood (ML) mixture}, is as follows:
\begin{align}
\label{eqn:hybrid}
    p(X_t|x^{t-1})=p(X_t|\phi(x^{t-1}))=\int_\theta p(X_t|\theta)p(\theta|\phi(x^{t-1}))
\end{align}
where $\phi(x^{t})$ is learned to maximize the following objective:
\begin{align}
  \mathcal{L}^\mathrm{MLM}_t(\phi)=&\log p(x^t|\phi)= 
  \sum_{i=1}^t \log \int_\theta p(x_i|\theta)p(\theta|\phi) \nonumber \\
  =&\sum_{i=1}^t \int_\theta q_t^*(\theta|x_i)\log\frac{p(x_i|\theta)p(\theta|\phi)}{q_t^*(\theta|x_i)} \nonumber \\
  \geq&\mathcal{L}^\mathrm{MLM}_{t-1}+\int_\theta q_t(\theta|x_t)\log \frac{p(x_t|\theta)p(\theta|\phi)}{q_t(\theta|x_t)} \nonumber \\
  &-\sum_{i=1}^{t-1}\int_\theta q_{t-1}(\theta|x_i)\log \frac{p(\theta|\phi_{t-1})}{p(\theta|\phi)}\nonumber \\
  \approx&\mathcal{L}^\mathrm{MLM}_{t-1}+\int_\theta q_t(\theta|x_t)\log \frac{p(x_t|\theta)p(\theta|\phi)}{q_t(\theta|x_t)} \nonumber \\ 
  &-(t-1)D_{\mathrm{KL}}[p(\theta|\phi_{t-1})\|p(\theta|\phi)]
  \label{eqn:ml_mixture}
\end{align}

$\phi_{t-1}$ minimizes $\mathcal{L}^\mathrm{MLM}_{t-1}(\phi)$ and $q^*_t$ denotes the theoretical optimal posterior that maximizes the ELBO, which means $q^*_t(\theta|x_i)=p(\theta|x_i,\phi_t)$. The approximation in (\ref{eqn:ml_mixture}) is based on the assumption that the previous objective $\mathcal{L}^\mathrm{MLM}_{t-1}$ is optimized good enough such that
\begin{equation}
    \label{eqn:mlm_approximation}
    p(\theta|\phi_{t-1})\approx\frac{1}{(t-1)}\sum_i^{t-1}q_{t-1}(\theta|x_i)
\end{equation}

For continual learning, in order to remove any dependency on previous data or previous approximate posterior, we optimize (\ref{eqn:ml_mixture}) sequentially in a greedy manner, namely, only the last two terms of (\ref{eqn:ml_mixture}) are optimized at time $t$:
\begin{align}
    \label{eqn:mlm_update}
    \phi_t=&\arg\max_\phi \int_\theta q_t(\theta|x_t)\log \frac{p(x_t|\theta)p(\theta|\phi)}{q_t(\theta|x_t)}\nonumber \\
    &-(t-1)D_{\mathrm{KL}}[p(\theta|\phi_{t-1})\|p(\theta|\phi)]
\end{align}

Like VCL, the optimization involved in ML Mixture code takes the form of a KL regularization in the parameter space. It addresses the two above-mentioned limitations of VCL. First, the prior overhead described in Sec. \ref{sec:overhead} is resolved because the mixture distribution is now optimized via maximum likelihood. Second, the approximation in (\ref{eqn:mlm_approximation}) is more plausible than (\ref{eqn:vcl_approximation}), because (\ref{eqn:vcl_approximation}) tries to approximate a fixed multimodal distribution with a simple distribution, while (\ref{eqn:mlm_approximation}) has optimizable components on both sides of the approximation.    

The performance of ML Mixture code in terms of forgetting and compression are also presented in Fig.\ref{forgetting} and \ref{compression}. One can see that although both VCL and ML Mixture use diagonal Gaussian for the distribution of model parameters (corresponding to $q(\theta|x^{t-1})$ in VCL and $p(\theta|\phi)$ in ML Mixture), ML Mixture achieves much less forgetting and smaller prequential codelength.

\subsection{The Intuitive Interpretation}
The ML Mixture code is effectively performing ML plug-in coding with a VAE. Unlike normal VAEs, the latent code includes the entire parameter vector of the decoder. This means that the mapping from the latent code to the input space is predetermined by the decoder architecture. Intuitively, ML Mixture projects each data $x_i$ from the input space to a local posterior distribution over the parameter space through the encoder $q(\theta|x_i)$. It then uses the prior distribution $p(\theta|\phi)$ to model the projected data in the parameter space. For each $x_i$, there exists a large number of simple distributions $q(\theta|x_i)$ that can be used as the encoder. Potentially, one can select the encoder so that the mirrored data points in the parameter space obeys a simple distribution, i.e. $\frac{1}{n}\sum_i^n q(\theta|x_i)$ is a simple distribution. In this case, $p(\theta|\phi)$ which is used to model the mixture distribution can also be kept simple. One supporting evidence for this possibility is the success of normal VAE/WAE\citep{tolstikhin2017wasserstein}. The prior distribution over the joint vector of latent code and the decoder parameters of VAE/WAE is a diagonal Gaussian, if we see the deterministic parameters as Gaussian random variables with zero variance.

\subsection{Limitation and Future Improvements}
There are a few limitations of the ML Mixture code that need to be solved before it is practical for continual learning. First, as we greedily optimize only the last two terms in (\ref{eqn:ml_mixture}) in a sequential manner, the previous approximate posterior $q_{t-1}$ is fixed once learned. This locks the prior distribution to a local region biased to the early tasks.  As a result, ML Mixture code learns well and retains the knowledge from early experiences. However, the description length is only marginally improved at the later tasks (see Appendix \ref{sec:10plots}). A potential fix is to introduce certain level of generative replay to optimize the first term in (\ref{eqn:ml_mixture}). Second, both VCL and ML Mixture use approximate objectives to remove dependency on previous data. Although (\ref{eqn:mlm_approximation}) could be more plausible than (\ref{eqn:vcl_approximation}), the error introduced in the objective could still be arbitrarily large as pointed out earlier in Sec. \ref{sec:var_gap}. In the future, it would be preferable to develop an alternative objective for which the approximation error is bounded.

\bibliography{example_paper}

\begin{thebibliography}{29}
\providecommand{\natexlab}[1]{#1}
\providecommand{\url}[1]{\texttt{#1}}
\expandafter\ifx\csname urlstyle\endcsname\relax
  \providecommand{\doi}[1]{doi: #1}\else
  \providecommand{\doi}{doi: \begingroup \urlstyle{rm}\Url}\fi

\bibitem[Arora et~al.(2018)Arora, Ge, Neyshabur, and Zhang]{arora2018stronger}
Arora, S., Ge, R., Neyshabur, B., and Zhang, Y.
\newblock Stronger generalization bounds for deep nets via a compression
  approach.
\newblock \emph{arXiv preprint arXiv:1802.05296}, 2018.

\bibitem[Blier \& Ollivier(2018)Blier and Ollivier]{blier2018description}
Blier, L. and Ollivier, Y.
\newblock The description length of deep learning models.
\newblock In \emph{Advances in Neural Information Processing Systems}, pp.\
  2216--2226, 2018.

\bibitem[Chaitin(2002)]{chaitin2002intelligibility}
Chaitin, G.~J.
\newblock On the intelligibility of the universe and the notions of simplicity,
  complexity and irreducibility.
\newblock \emph{arXiv preprint math/0210035}, 2002.

\bibitem[Ebrahimi et~al.(2020)Ebrahimi, Elhoseiny, Darrell, and
  Rohrbach]{Ebrahimi2020Uncertainty-guided}
Ebrahimi, S., Elhoseiny, M., Darrell, T., and Rohrbach, M.
\newblock Uncertainty-guided continual learning with bayesian neural networks.
\newblock In \emph{International Conference on Learning Representations}, 2020.

\bibitem[Farquhar \& Gal(2018)Farquhar and Gal]{farquhar2018towards}
Farquhar, S. and Gal, Y.
\newblock Towards robust evaluations of continual learning.
\newblock \emph{arXiv preprint arXiv:1805.09733}, 2018.

\bibitem[Farquhar \& Gal(2019)Farquhar and Gal]{farquhar2019unifying}
Farquhar, S. and Gal, Y.
\newblock A unifying bayesian view of continual learning.
\newblock \emph{arXiv preprint arXiv:1902.06494}, 2019.

\bibitem[Gr{\"u}nwald(2007)]{grunwald2007minimum}
Gr{\"u}nwald, P.~D.
\newblock \emph{The minimum description length principle}.
\newblock The MIT Press, 2007.

\bibitem[Hinton \& Van~Camp(1993)Hinton and Van~Camp]{hinton1993keeping}
Hinton, G.~E. and Van~Camp, D.
\newblock Keeping the neural networks simple by minimizing the description
  length of the weights.
\newblock In \emph{Proceedings of the sixth annual conference on Computational
  learning theory}, pp.\  5--13, 1993.

\bibitem[Javed \& White(2019)Javed and White]{javed2019meta}
Javed, K. and White, M.
\newblock Meta-learning representations for continual learning.
\newblock In \emph{Advances in Neural Information Processing Systems}, pp.\
  1818--1828, 2019.

\bibitem[Kingma \& Welling(2013)Kingma and Welling]{kingma2013auto}
Kingma, D.~P. and Welling, M.
\newblock Auto-encoding variational bayes.
\newblock \emph{arXiv preprint arXiv:1312.6114}, 2013.

\bibitem[Kirkpatrick et~al.(2017)Kirkpatrick, Pascanu, Rabinowitz, Veness,
  Desjardins, Rusu, Milan, Quan, Ramalho, Grabska-Barwinska,
  et~al.]{kirkpatrick2017overcoming}
Kirkpatrick, J., Pascanu, R., Rabinowitz, N., Veness, J., Desjardins, G., Rusu,
  A.~A., Milan, K., Quan, J., Ramalho, T., Grabska-Barwinska, A., et~al.
\newblock Overcoming catastrophic forgetting in neural networks.
\newblock \emph{Proceedings of the national academy of sciences}, 114\penalty0
  (13):\penalty0 3521--3526, 2017.

\bibitem[Kolmogorov(1968)]{kolmogorov1968three}
Kolmogorov, A.~N.
\newblock Three approaches to the quantitative definition of information.
\newblock \emph{International journal of computer mathematics}, 2\penalty0
  (1-4):\penalty0 157--168, 1968.

\bibitem[Krueger et~al.(2017)Krueger, Huang, Islam, Turner, Lacoste, and
  Courville]{krueger2017bayesian}
Krueger, D., Huang, C.-W., Islam, R., Turner, R., Lacoste, A., and Courville,
  A.
\newblock Bayesian hypernetworks.
\newblock \emph{arXiv preprint arXiv:1710.04759}, 2017.

\bibitem[Lesort et~al.(2019)Lesort, Caselles-Dupr{\'e}, Garcia-Ortiz, Stoian,
  and Filliat]{lesort2019generative}
Lesort, T., Caselles-Dupr{\'e}, H., Garcia-Ortiz, M., Stoian, A., and Filliat,
  D.
\newblock Generative models from the perspective of continual learning.
\newblock In \emph{2019 International Joint Conference on Neural Networks
  (IJCNN)}, pp.\  1--8. IEEE, 2019.

\bibitem[Lopez-Paz \& Ranzato(2017)Lopez-Paz and Ranzato]{lopez2017gradient}
Lopez-Paz, D. and Ranzato, M.
\newblock Gradient episodic memory for continual learning.
\newblock In \emph{Advances in Neural Information Processing Systems}, pp.\
  6467--6476, 2017.

\bibitem[McCloskey \& Cohen(1989)McCloskey and
  Cohen]{mccloskey1989catastrophic}
McCloskey, M. and Cohen, N.~J.
\newblock Catastrophic interference in connectionist networks: The sequential
  learning problem.
\newblock In \emph{Psychology of learning and motivation}, volume~24, pp.\
  109--165. Elsevier, 1989.

\bibitem[Nguyen et~al.(2018)Nguyen, Li, Bui, and Turner]{2018variational}
Nguyen, C.~V., Li, Y., Bui, T.~D., and Turner, R.~E.
\newblock Variational continual learning.
\newblock In \emph{International Conference on Learning Representations}, 2018.

\bibitem[Ollivier(2014)]{ollivier2014auto}
Ollivier, Y.
\newblock Auto-encoders: reconstruction versus compression.
\newblock \emph{arXiv preprint arXiv:1403.7752}, 2014.

\bibitem[Opper(1998)]{opper1998bayesian}
Opper, M.
\newblock A bayesian approach to on-line learning.
\newblock \emph{On-line learning in neural networks}, pp.\  363--378, 1998.

\bibitem[Parisi et~al.(2019)Parisi, Kemker, Part, Kanan, and
  Wermter]{parisi2019continual}
Parisi, G.~I., Kemker, R., Part, J.~L., Kanan, C., and Wermter, S.
\newblock Continual lifelong learning with neural networks: A review.
\newblock \emph{Neural Networks}, 2019.

\bibitem[Rao et~al.(2019)Rao, Visin, Rusu, Pascanu, Teh, and
  Hadsell]{rao2019continual}
Rao, D., Visin, F., Rusu, A., Pascanu, R., Teh, Y.~W., and Hadsell, R.
\newblock Continual unsupervised representation learning.
\newblock In \emph{Advances in Neural Information Processing Systems}, pp.\
  7645--7655, 2019.

\bibitem[Ring(1998)]{ring1998child}
Ring, M.~B.
\newblock Child: A first step towards continual learning.
\newblock In \emph{Learning to learn}, pp.\  261--292. Springer, 1998.

\bibitem[Ritter et~al.(2018)Ritter, Botev, and Barber]{ritter2018online}
Ritter, H., Botev, A., and Barber, D.
\newblock Online structured laplace approximations for overcoming catastrophic
  forgetting.
\newblock In \emph{Advances in Neural Information Processing Systems}, pp.\
  3738--3748, 2018.

\bibitem[Shin et~al.(2017)Shin, Lee, Kim, and Kim]{shin2017continual}
Shin, H., Lee, J.~K., Kim, J., and Kim, J.
\newblock Continual learning with deep generative replay.
\newblock In \emph{Advances in Neural Information Processing Systems}, pp.\
  2990--2999, 2017.

\bibitem[Solomonoff(1964)]{solomonoff1964formal}
Solomonoff, R.~J.
\newblock A formal theory of inductive inference. part i.
\newblock \emph{Information and control}, 7\penalty0 (1):\penalty0 1--22, 1964.

\bibitem[Swaroop et~al.(2019)Swaroop, Nguyen, Bui, and
  Turner]{swaroop2019improving}
Swaroop, S., Nguyen, C.~V., Bui, T.~D., and Turner, R.~E.
\newblock Improving and understanding variational continual learning.
\newblock \emph{arXiv preprint arXiv:1905.02099}, 2019.

\bibitem[Tishby \& Zaslavsky(2015)Tishby and Zaslavsky]{tishby2015deep}
Tishby, N. and Zaslavsky, N.
\newblock Deep learning and the information bottleneck principle.
\newblock In \emph{2015 IEEE Information Theory Workshop (ITW)}, pp.\  1--5.
  IEEE, 2015.

\bibitem[Tolstikhin et~al.(2017)Tolstikhin, Bousquet, Gelly, and
  Schoelkopf]{tolstikhin2017wasserstein}
Tolstikhin, I., Bousquet, O., Gelly, S., and Schoelkopf, B.
\newblock Wasserstein auto-encoders.
\newblock \emph{arXiv preprint arXiv:1711.01558}, 2017.

\bibitem[Wu et~al.(2018)Wu, Herranz, Liu, wang, van~de Weijer, and
  Raducanu]{NIPS2018_7836}
Wu, C., Herranz, L., Liu, X., wang, y., van~de Weijer, J., and Raducanu, B.
\newblock Memory replay gans: Learning to generate new categories without
  forgetting.
\newblock In Bengio, S., Wallach, H., Larochelle, H., Grauman, K.,
  Cesa-Bianchi, N., and Garnett, R. (eds.), \emph{Advances in Neural
  Information Processing Systems 31}, pp.\  5962--5972. Curran Associates,
  Inc., 2018.

\end{thebibliography}
\bibliographystyle{icml2020}

\newpage
\newpage
\appendix 
\section{Description length on each digit through out training}
\label{sec:10plots}
\begin{figure}[h!]
\label{fig:perclass_bpd}
    \begin{subfigure}[c]{.45\linewidth}
    \includegraphics[width=\linewidth]{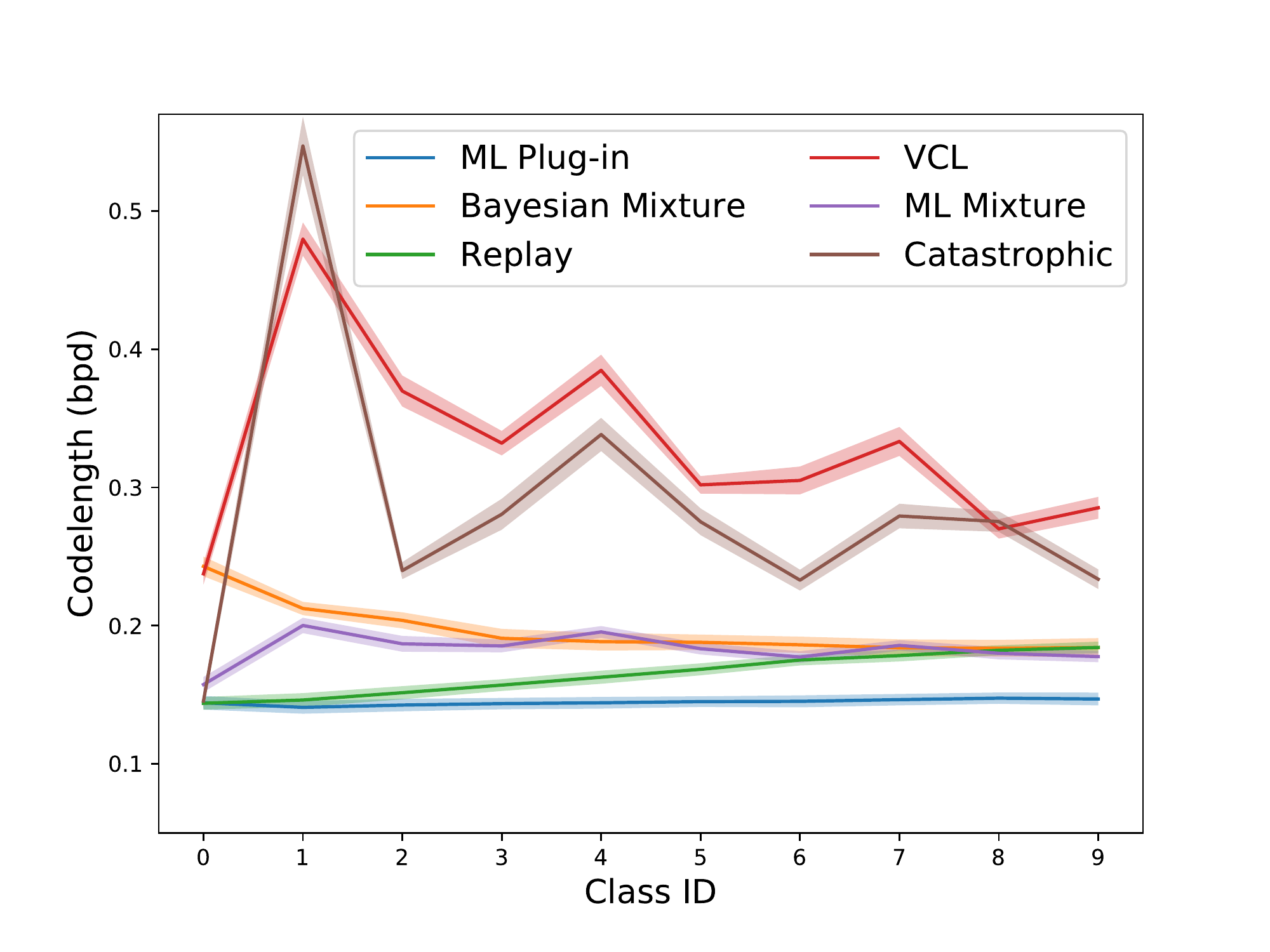}
    \caption{0}
    \end{subfigure}\hfill
    \begin{subfigure}[c]{.45\linewidth}
    \includegraphics[width=\linewidth]{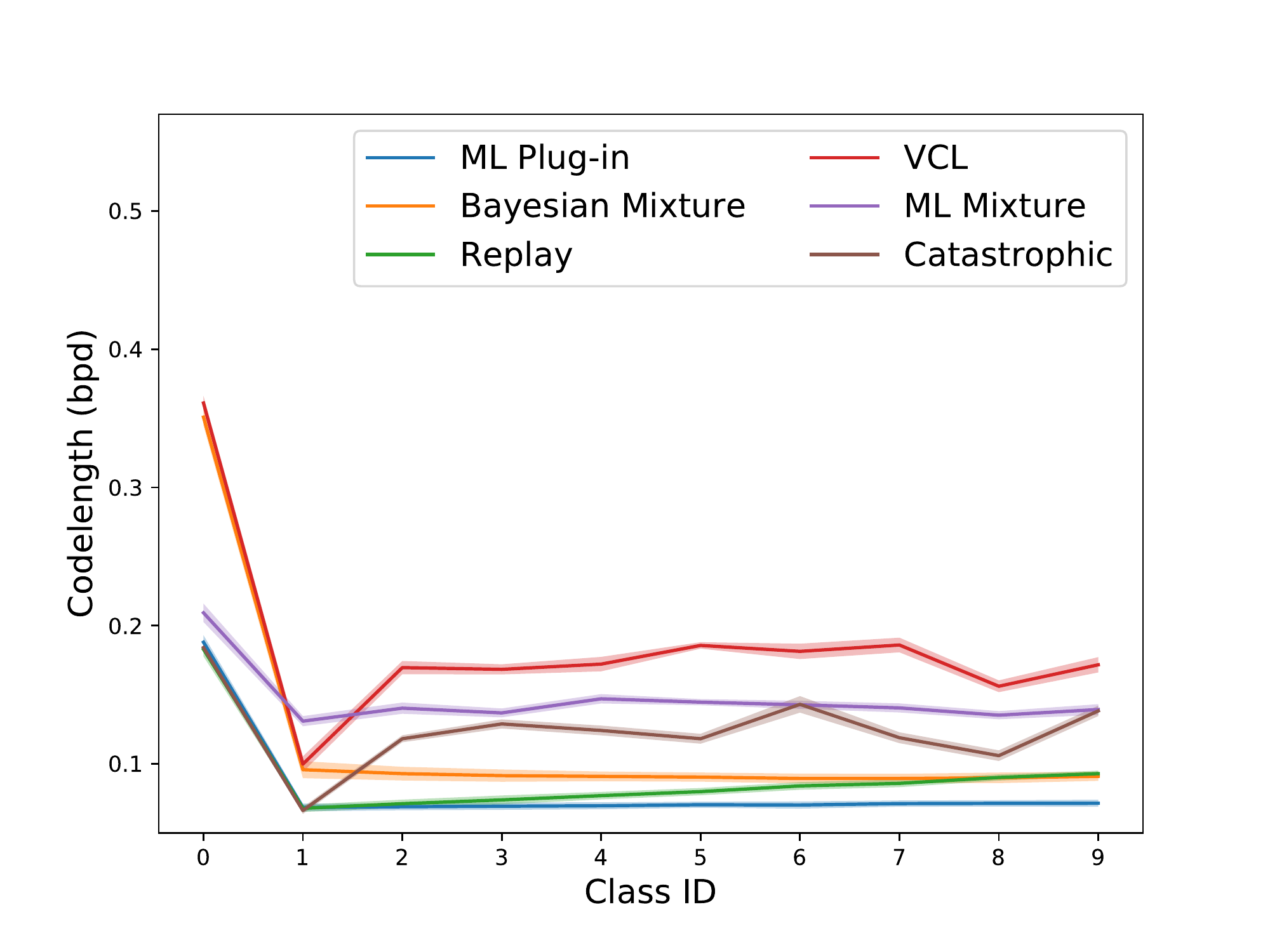}
    \caption{1}
    \end{subfigure}
    \begin{subfigure}[c]{.45\linewidth}
    \includegraphics[width=\linewidth]{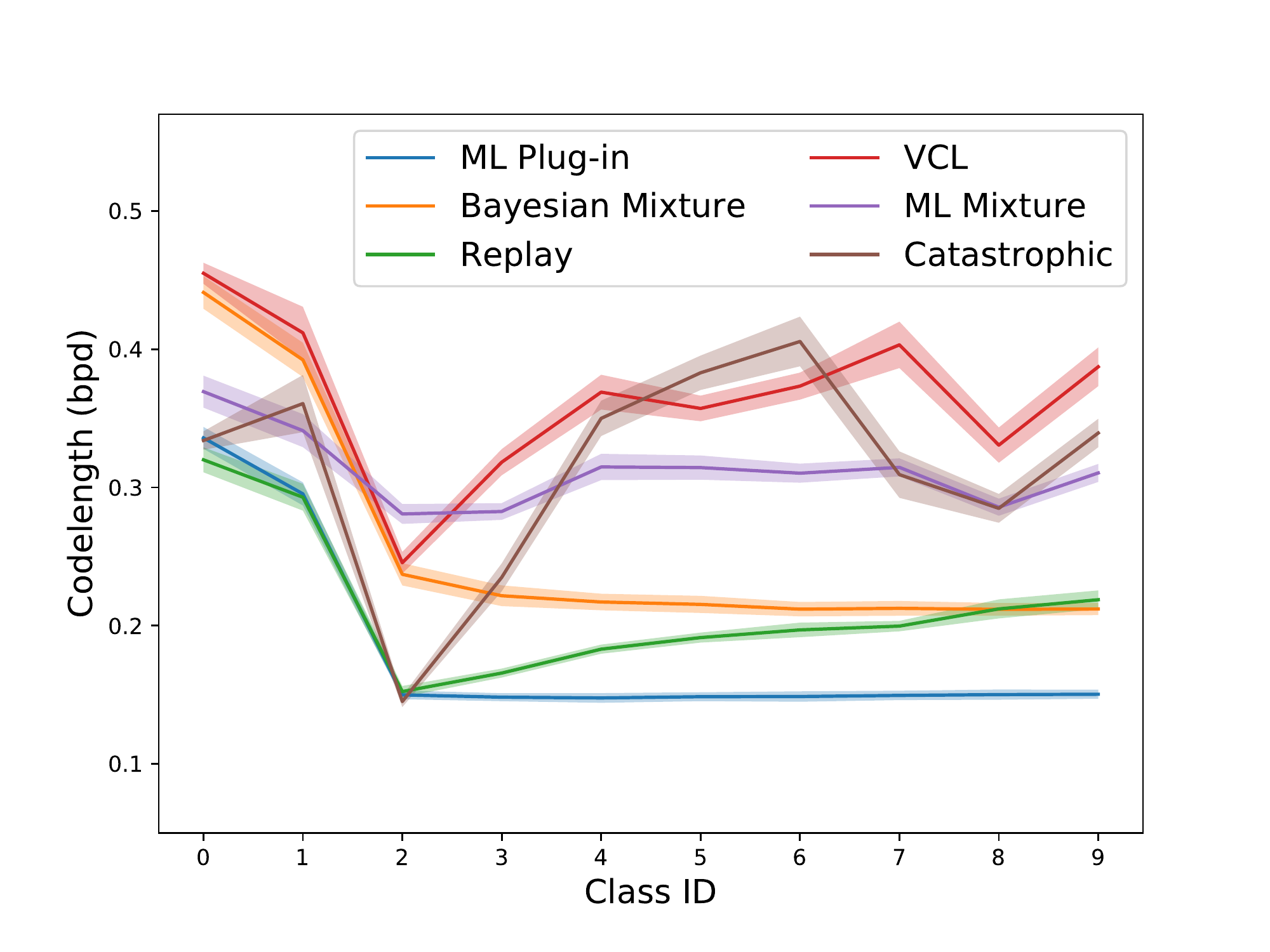}
    \caption{2}
    \end{subfigure}
    \begin{subfigure}[c]{.45\linewidth}
    \includegraphics[width=\linewidth]{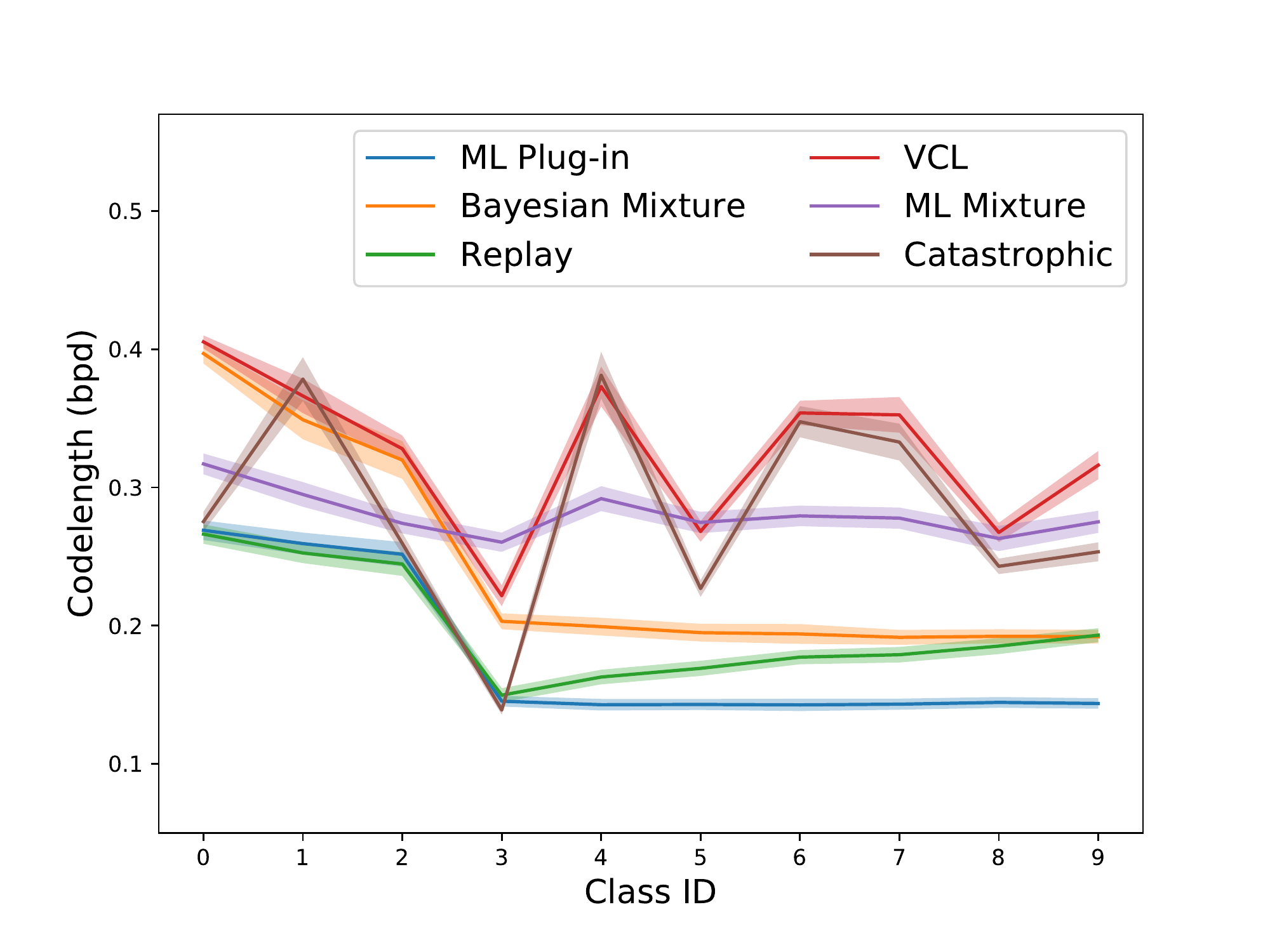}
    \caption{3}
    \end{subfigure}
    \begin{subfigure}[c]{.45\linewidth}
    \includegraphics[width=\linewidth]{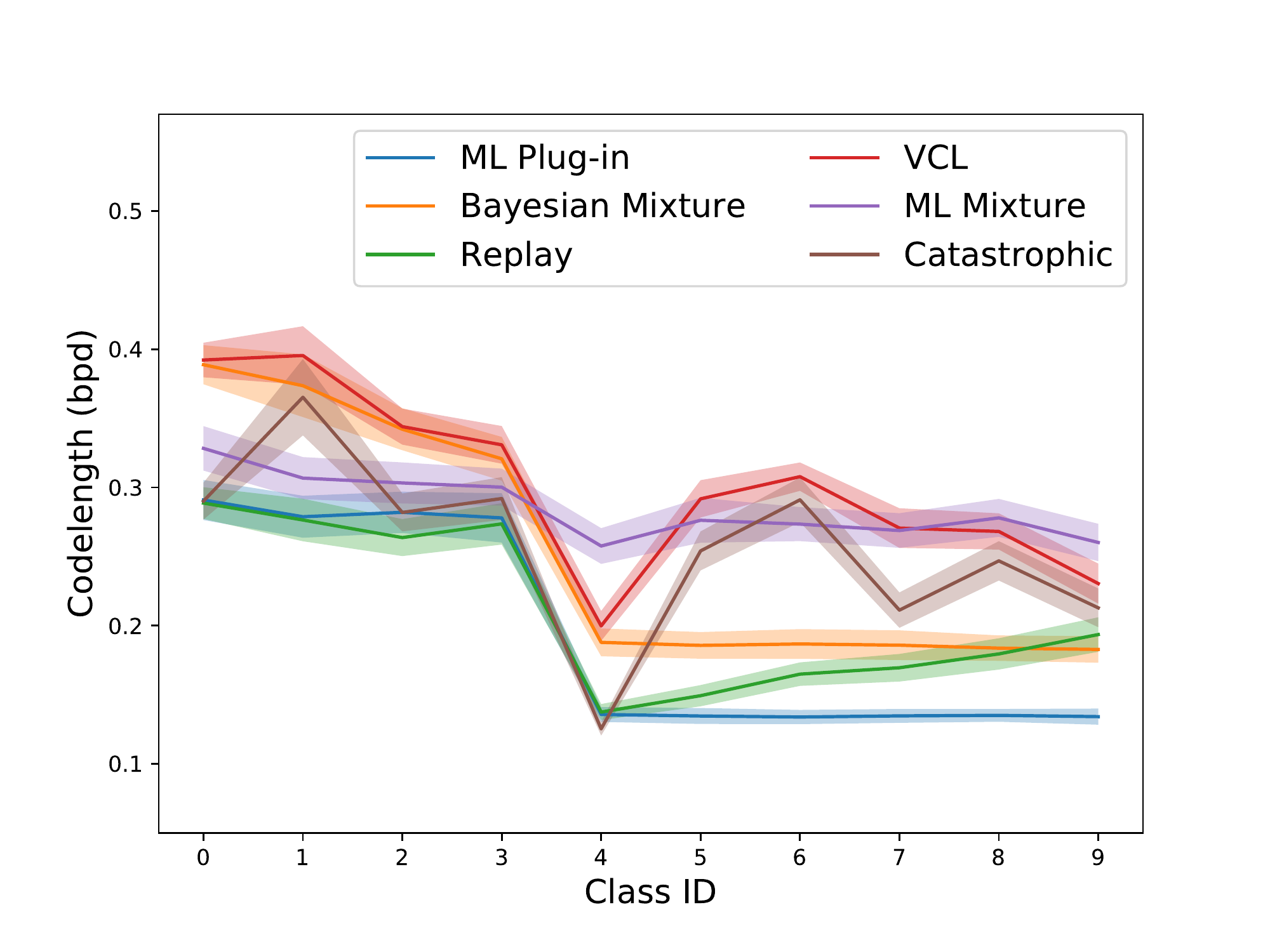}
    \caption{4}
    \end{subfigure}
    \begin{subfigure}[c]{.45\linewidth}
    \includegraphics[width=\linewidth]{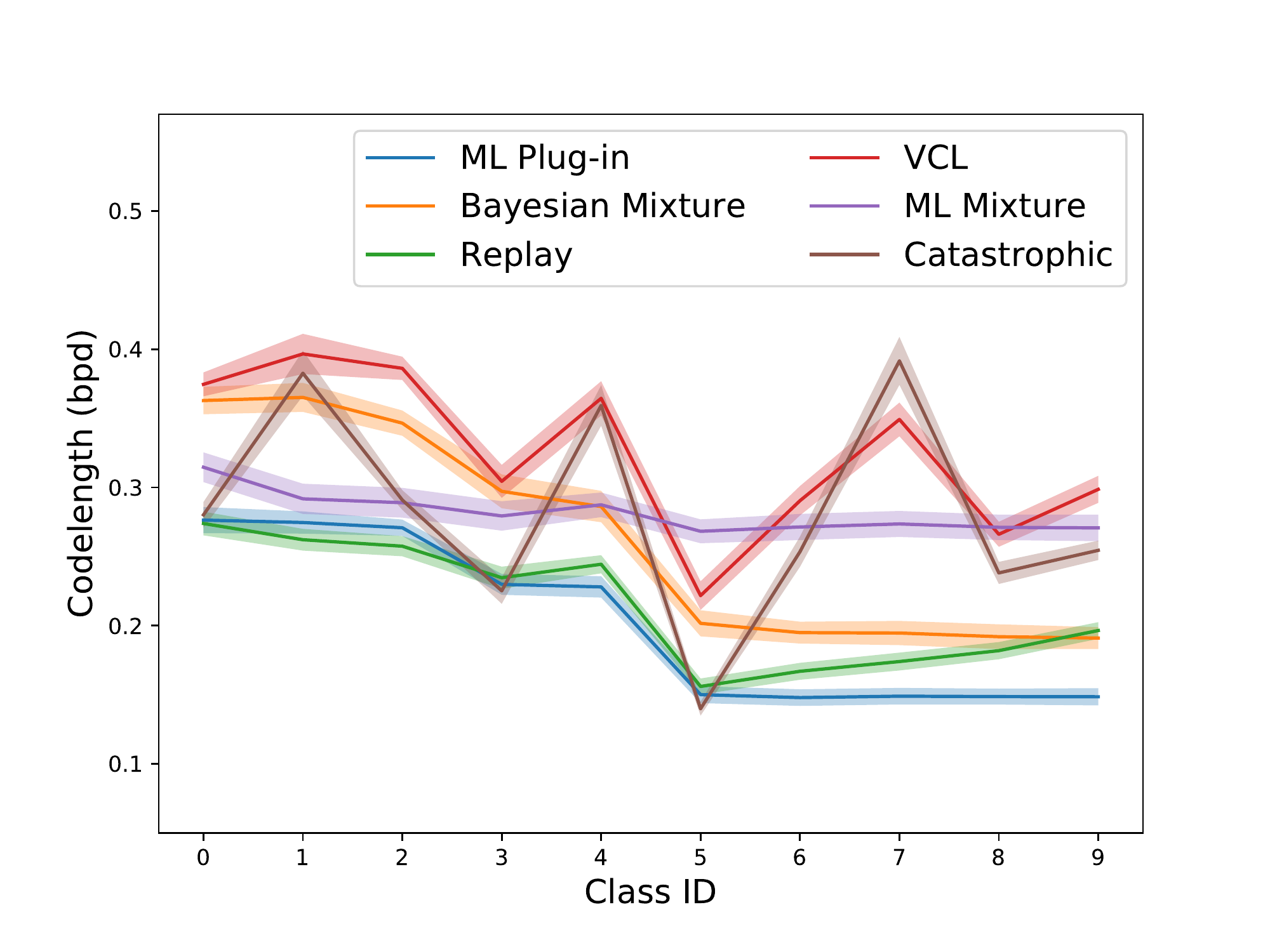}
    \caption{5}
    \end{subfigure}
    \begin{subfigure}[c]{.45\linewidth}
    \includegraphics[width=\linewidth]{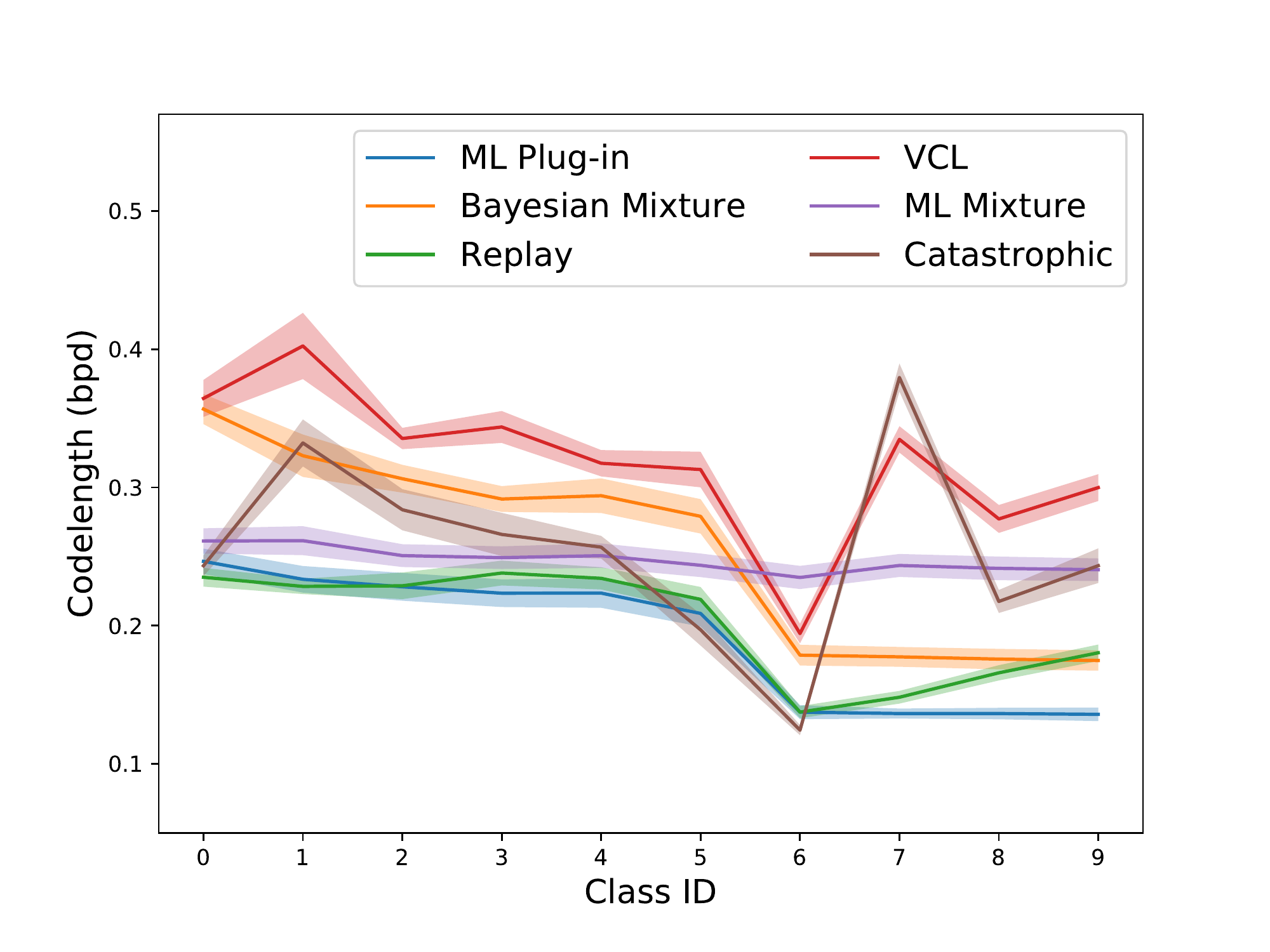}
    \caption{6}
    \end{subfigure}
    \begin{subfigure}[c]{.45\linewidth}
    \includegraphics[width=\linewidth]{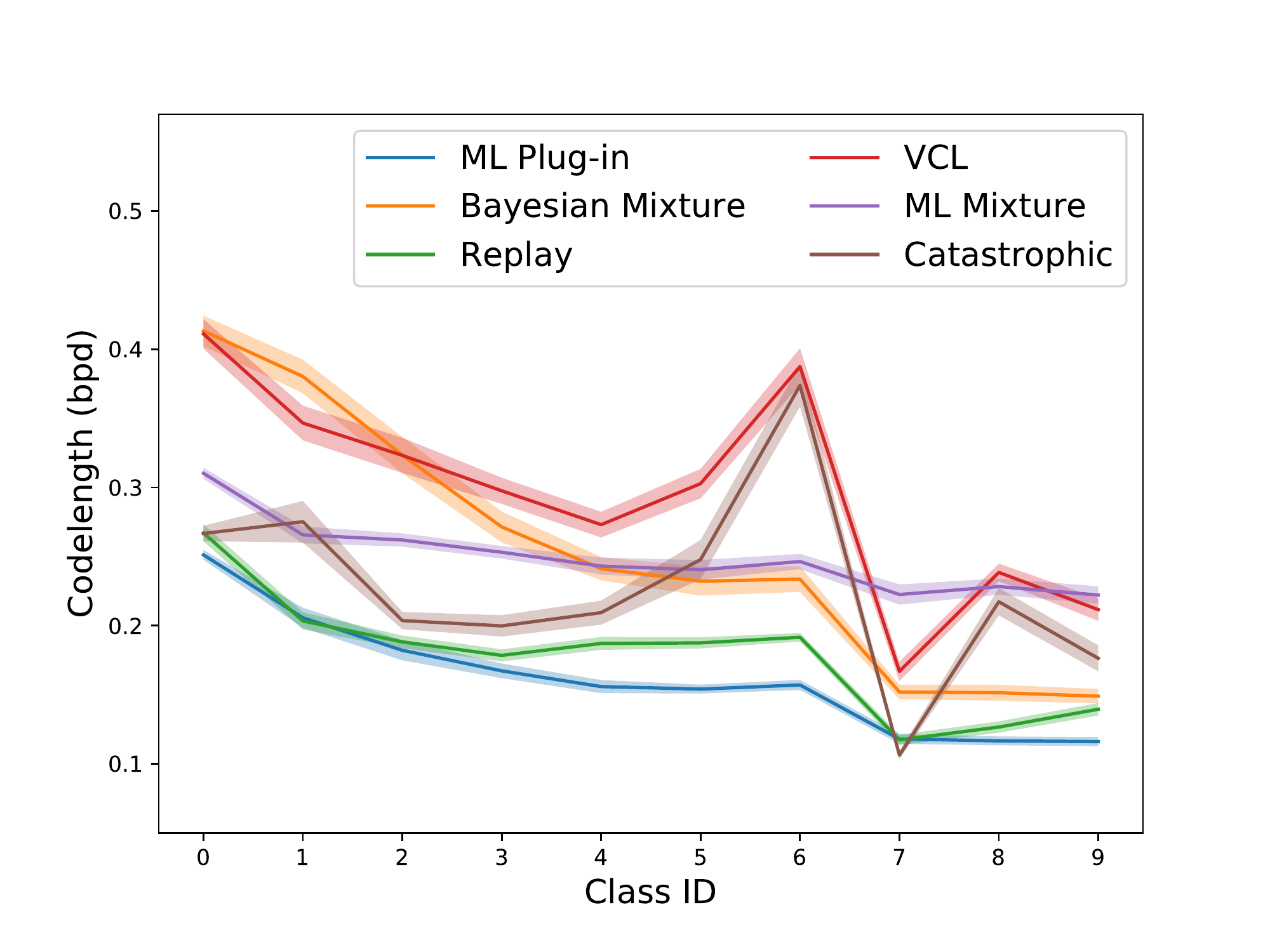}
    \caption{7}
    \end{subfigure}
    \begin{subfigure}[c]{.45\linewidth}
    \includegraphics[width=\linewidth]{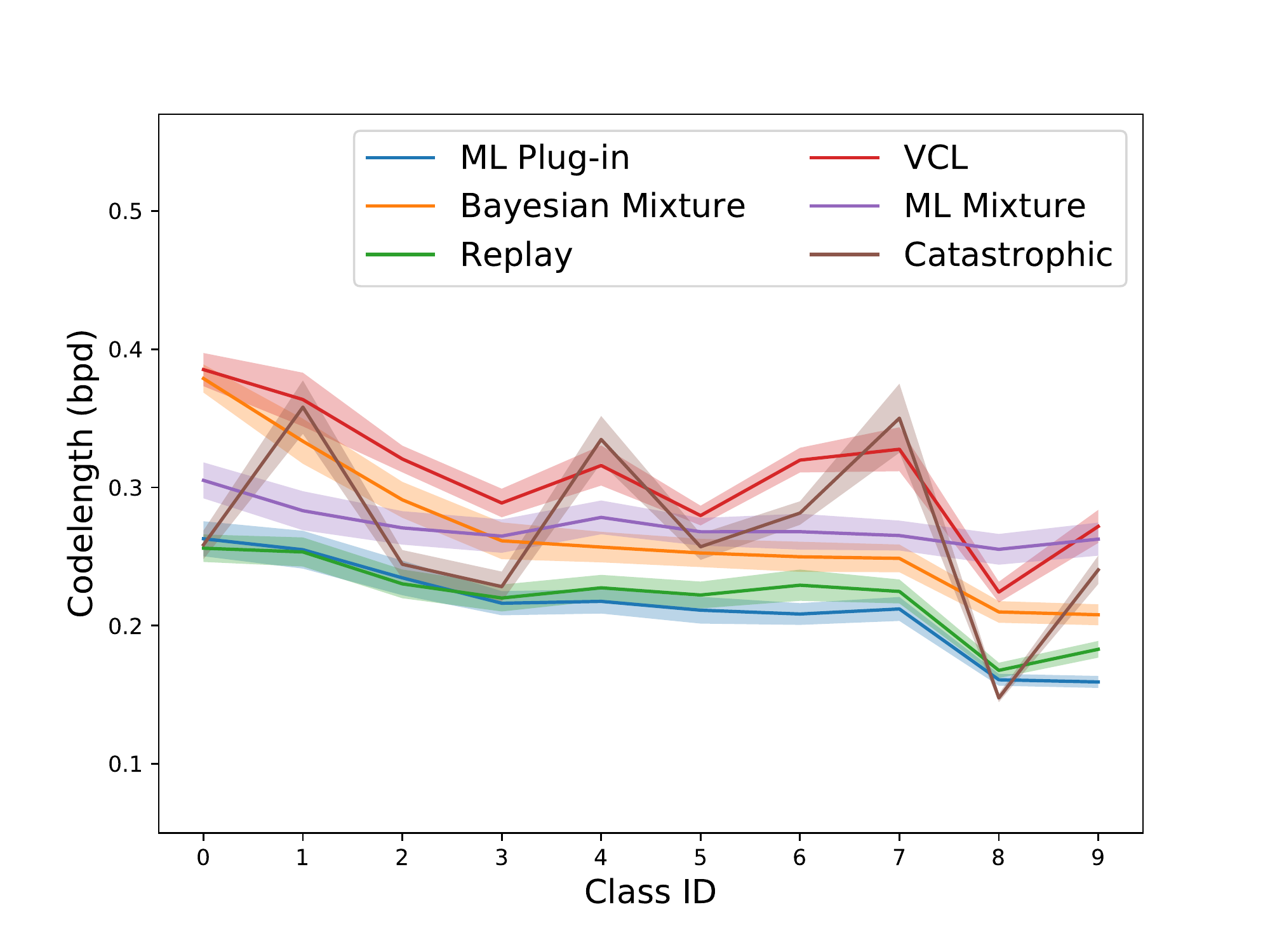}
    \caption{8}
    \end{subfigure}\hfill
    \begin{subfigure}[c]{.45\linewidth}
    \includegraphics[width=\linewidth]{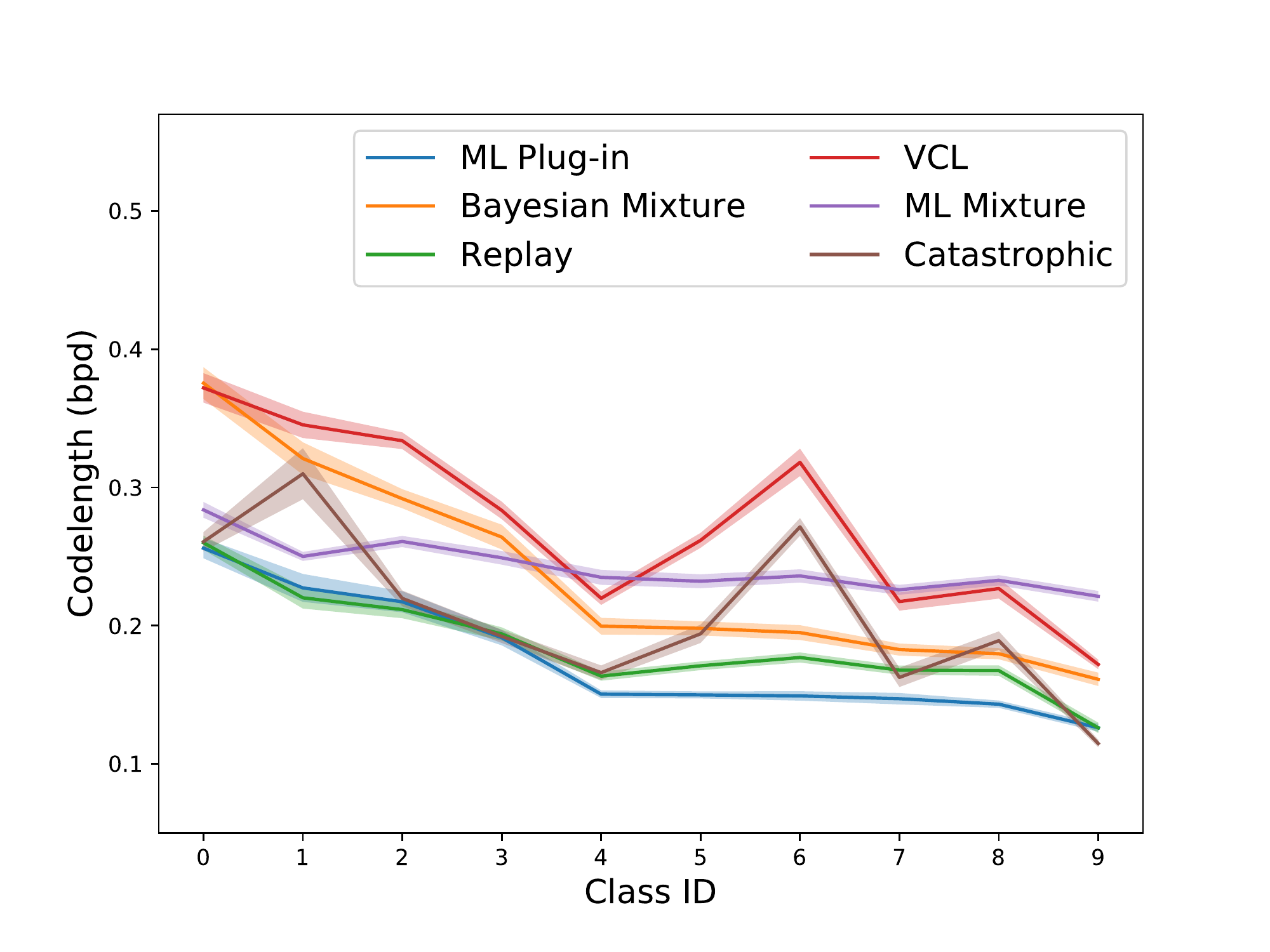}
    \caption{9}
    \end{subfigure}
    \caption{The change of description length on different class through out training.}
    
\end{figure}

\end{document}